\pdfoutput=1
\documentclass[11pt]{article}

\usepackage[final]{acl}

\usepackage{times}
\usepackage{latexsym}

\usepackage[T1]{fontenc}
\usepackage[utf8]{inputenc}

\usepackage{microtype}

\usepackage{inconsolata}

\usepackage{graphicx}

\usepackage{natbib}
\usepackage{doi}
\usepackage{algorithm}
\usepackage{algpseudocode}
\usepackage{amsmath}
\usepackage{booktabs}  % 引入 booktabs 宏包
\usepackage{paralist}
\usepackage{amsthm}
\usepackage{multirow}
\usepackage{listings}

\newtheorem{definition}{Definition}

\title{Simple Yet Effective: Extracting Private Data Across Clients in Federated Fine-Tuning of Large Language Models}
\author{
\textbf{Yingqi Hu\textsuperscript{1}}, 
\textbf{Zhuo Zhang\textsuperscript{1}}, 
\textbf{Jingyuan Zhang\textsuperscript{2}}, 
\textbf{Jinghua Wang\textsuperscript{1}},\\
{\bf
\textbf{Qifan Wang\textsuperscript{3}}, 
\textbf{Lizhen Qu\textsuperscript{4}\thanks{Corresponding author.}}, 
\textbf{Zenglin Xu\textsuperscript{5,6}}
}\\
\\
\textsuperscript{1}Harbin Institute of Technology, Shenzhen, China\\
\textsuperscript{2}Kuaishou Technology, China \quad
\textsuperscript{3}Meta AI, USA\\
\textsuperscript{4}Monash University, Australia \quad
\textsuperscript{5}Fudan University, China\\
\textsuperscript{6}Shanghai Academy of Artificial Intelligence for Science, China
}

\begin{document}
\maketitle
\begin{abstract}
Federated large language models (FedLLMs) enable cross-silo collaborative training among institutions while preserving data locality, making them appealing for privacy-sensitive domains such as law, finance, and healthcare. However, the memorization behavior of LLMs can lead to privacy risks that may cause cross-client data leakage. In this work, we study the threat of \emph{cross-client data extraction}, where a semi-honest participant attempts to recover personally identifiable information (PII) memorized from other clients’ data. We propose three simple yet effective extraction strategies that leverage contextual prefixes from the attacker’s local data, including frequency-based prefix sampling and local fine-tuning to amplify memorization. To evaluate these attacks, we construct a Chinese legal-domain dataset with fine-grained PII annotations consistent with CPIS, GDPR, and CCPA standards, and assess extraction performance using two metrics: \emph{coverage} and \emph{efficiency}. Experimental results show that our methods can recover up to 56.6\% of victim-exclusive PII, where names, addresses, and birthdays are particularly vulnerable. These findings highlight concrete privacy risks in FedLLMs and establish a benchmark and evaluation framework for future research on privacy-preserving federated learning. Code and data are available at \href{https://github.com/SMILELab-FL/FedPII}{https://github.com/SMILELab-FL/FedPII}.
\end{abstract}

\section{Introduction}
\label{sec:introduction}

Federated large language models (FedLLMs)~\citep{fedllm_benchmark_NEURIPS2024, ye2024openfedllm, fedlegal_acl_2023, Patterns2024_fedllm_position, survey_yao2024federatedlargelanguagemodels} have recently emerged as a promising approach for cross-silo federated learning (FL)~\citep{li2024fedcompass}, enabling collaborative model training while maintaining data locality and institutional privacy. In cross-silo FL, organizations, such as courts, banks, and hospitals, collaboratively fine-tune\footnote{In current practice, \emph{FedLLM} typically refers to the federated fine-tuning of large language models rather than federated pre-training. See Appendix~\ref{app:background_knowledge_FedLLMs} for details.} a shared model without exchanging private records. Prior studies mainly focused on improving algorithmic efficiency and convergence~\citep{fedfinetune_NEURIPS2024_1a134b50, survey_fed_LLM_wu2025surveyfederatedfinetuninglarge, fedprox_2020}, while the privacy vulnerabilities of FedLLMs remain largely unexplored.

A growing body of work has shown that large language models (LLMs) tend to memorize and reproduce fragments of their training data, including PII such as names, addresses, and dates of birth~\citep{carlini2021extractingtrainingdatalarge, quantifying_iclr_2023, associative_memorization_shao-etal-2024-quantifying, spt_pii_nips_2023, benchmark_nakka2024piiscopebenchmarktrainingdata}. Although FL mitigates privacy risks by exchanging model updates instead of raw data, our preliminary experiments (Appendix~\ref{sec:prelim_verbatim_attack}) indicate that FedLLMs remain vulnerable to \emph{verbatim data extraction} (VDE)—where adversaries can recover verbatim text sequences from the aggregated global model. However, most existing VDE studies assume that attackers have privileged access or significant knowledge of the victim’s data~\citep{bag_of_tricks_icml_2023, llm_leakage_question_2022huang_etal}, which is unrealistic in practical cross-silo deployments.

\begin{figure*}[t]
\centering
\includegraphics[width=1\linewidth]{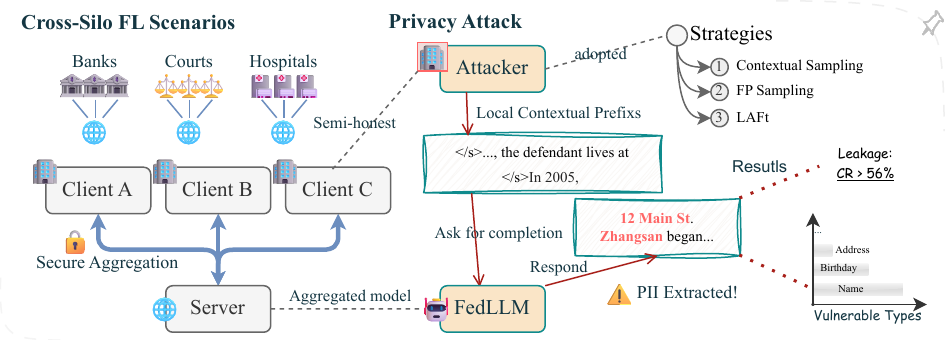}
\caption{Overview of cross-silo FedLLMs and the proposed privacy attack.
In cross-silo FL, institutions such as banks, courts, and hospitals collaboratively fine-tune a shared model under the coordination of a central server, keeping data local.
A semi-honest client leverages its local data to construct PII-related prefixes and queries the aggregated global FedLLM, leading to cross-client data leakage.
The proposed strategies—Contextual Prefix Sampling, Frequency-prioritized (FP) Sampling, and Latent Association Fine-tuning (LAFt)—achieve up to 56.6\% recovery of victim-exclusive PII, with names, addresses, and birthdays being the most vulnerable categories.}
\label{fig:overview}
\end{figure*}

We instead consider a more realistic \emph{semi-honest} threat model, where each participant follows the FL protocol but may attempt to infer private information from the global model. For instance, in a federation of courts, a participant could exploit its own local case records as contextual prefixes to elicit sensitive information memorized from other courts’ data (as illustrated in Figure~\ref{fig:overview}). To investigate this, we propose three extraction strategies:  
(1) PII Contextual Prefix Sampling, which queries the global model using local contextual prefixes;  
(2) Frequency-prioritized (FP) Sampling, which focuses on high-frequency prefixes to improve extraction efficiency; and  
(3) Latent Association Fine-tuning (LAFt), which locally fine-tunes the global model using stored prefix-PII pairs, thereby improving its ability to extract PIIs implicitly memorized within the model.%locally fine-tunes the global model on prefix–PII pairs to enhance extraction from the memory of the model.

To evaluate these attacks, we build a benchmark dataset by annotating a real-world Chinese legal corpus with fine-grained PII labels aligned with privacy regulations such as CPIS, GDPR, and CCPA (see Acronyms List~\ref{sec:acronyms}). We assess the attacks using two metrics: \emph{coverage} (the proportion of target PII successfully extracted) and \emph{efficiency} (the amount of PII recovered under a limited query budget). Our experiments reveal that the proposed attacks can extract up to 56.6\% of victim-exclusive PII, with \textit{Name}, \textit{Address}, and \textit{Birthday} being the most vulnerable types. Moreover, we observe diminishing returns as query budgets grow, while FP Sampling and LAFt enhance diversity under tighter budgets. These findings expose concrete privacy risks in FedLLMs and underscore the need for stronger privacy-preserving mechanisms.

In summary, our main contributions are:
\begin{compactenum}
    \item We propose three novel extraction strategies for FedLLMs, independent of existing gradient-based or membership inference attacks, and evaluate them using two rigorous metrics: coverage and efficiency.
    \item Extensive experiments show that our methods can recover up to 56.6\% of cross-client unique PII, with larger prefix sets yielding diminishing returns in efficiency, revealing a trade-off between coverage and computational cost.
    \item We construct a real-world benchmark dataset by augmenting a legal-domain corpus with fine-grained PII annotations aligned with CPIS, GDPR, and CCPA standards, filling the gap in public resources for privacy research in federated learning.
\end{compactenum}

\section{Related Work}
This study is related to the fields of data extraction attacks and federated learning. For the reader's convenience, a brief introduction to these concepts is provided in Appendix~\ref{sec:preliminary}. In this section, we review only the work directly related to our method.

\subsection{PII Extraction Attacks in LLM}
Large language models, due to their massive parameter scale, are capable of memorizing exact training data samples, making them vulnerable to data extraction attacks. These attacks can target different granularities of information: sample-level and entity-level.

At the sample level, an attacker with access to the full prefix of a training sample can query the LLM to regenerate the exact suffix~\citep{bag_of_tricks_icml_2023, min_k_iclr_2024, min_k_pp_iclr_2025}. This technique, known as \textit{verbatim training data extraction}~\citep{carlini2021extractingtrainingdatalarge, quantifying_iclr_2023, extraction_lens_nips2024}, is widely used to detect data contamination and copyright violations~\citep{CDD_acl_2024}.

At the entity level, attackers may know a subset of PII entities—such as names or affiliations—about a particular subject. By combining these known details with prompt templates (either manually crafted or automatically generated~\citep{kassem2025alpacavicunausingllms}), they can elicit the model to produce additional PII records about the same subject. This is known as an associative data extraction attack~\citep{associative_memorization_shao-etal-2024-quantifying, spt_pii_nips_2023, entity_extract_aaai_2024}.

Broadly, PII extraction attacks refer to any attack that aims at eliciting outputs from the model that contain PII~\citep{PII_leakage_2023_IEEE_SP, benchmark_nakka2024piiscopebenchmarktrainingdata, llm_leakage_question_2022huang_etal}. Both verbatim and associative techniques can be used to conduct such attacks.

While most prior work assumes centralized training with full data access, we investigate PII extraction under federated fine-tuning, where the attacker has limited observability and control. We elaborate on this in Section~\ref{sec:prob_def}.

% \vspace{-6pt}

\subsection{Privacy Threats in Federated Learning}
Threats in Federated Learning can be categorized into two main areas: security and privacy~\citep{survey_linkagesecurityprivacyfairness_2024wang, survey_XIE2024127225, survey_li2024synergizingfoundationmodelsfederated}. Security threats typically aim to disrupt the entire FL system by invalidating model training~\citep{shejwalkar2021manipulating} and introducing backdoors~\citep{backdoor_pmlr-v108-bagdasaryan20a, backdoor_fedtrojan_2024}. In contrast, privacy threats have attracted more attention from researchers and focus on stealing confidential information from the FL system, such as inferring sensitive properties~\citep{attribute_inference_8835269}, reconstructing clients' private datasets~\citep{NIPS2019_DLG, invert_grads_2020}, and determining the membership and source of training data~\citep{fltrojan_mia_rashid2025, fedllm_mia_vu2024analysisprivacyleakagefederated, source_sia_mia}. To achieve these attacks, researchers often make different assumptions regarding the attacker's knowledge. Common assumptions typically fall into two dimensions: whether the attacker is a client or a server~\citep{panning_for_gold_server_side_ChuGFGG23}, and whether the attacker is semi-honest~\citep{Applebaum2017, source_sia_mia} or malicious. These assumptions determine whether the attacker has access to gradients, local datasets, model parameters, and the ability to manipulate them.

\section{Dataset}
\label{sec:dataset}
\subsection{Data Sources and Preprocessing}

The majority of our dataset is sourced from the Challenge of AI in Law (CAIL)~\citep{cail_lexeval_nips_2024}, supplemented by smaller portions from CJRC~\citep{CJRC} and JEC-QA~\citep{JEC-QA}. CAIL is a renowned annual competition featuring a variety of legal NLP tasks. In this study, we focus on two natural language generation tasks: \textit{Judicial Summary} and \textit{Judicial Reading Comprehension}, and three natural language understanding tasks: \textit{Similar Case Matching}, \textit{Judicial Exam Question Answering}, and \textit{Legal Case Classification}.
Detailed task descriptions are provided in Appendix~\ref{sec:tasks_desc}, with representative examples shown in Table~\ref{tab:task_input_output_eg}.

Following prior work~\citep{fedlegal_acl_2023, yue2024lawllm}, we further preprocess and curate the dataset to fit our setting. The complete preprocessing pipeline is described in Appendix~\ref{sec:ds_preprocessing}, where Table~\ref{tab:dataset_stats} reports the dataset statistics.\footnote{The datasets contain PII from publicly available government-published legal documents. They were de-identified and used in prior work, e.g., as in \citet{yue2024lawllm}. We use curated versions from these papers. Since our study concerns privacy risks in FedLLMs, real-world PII is necessary to evaluate model vulnerabilities.}

\subsection{PII Labeling}

\begin{figure}[t]
    \centering
    % \hspace*{-15pt}
    \includegraphics[width=1\linewidth]{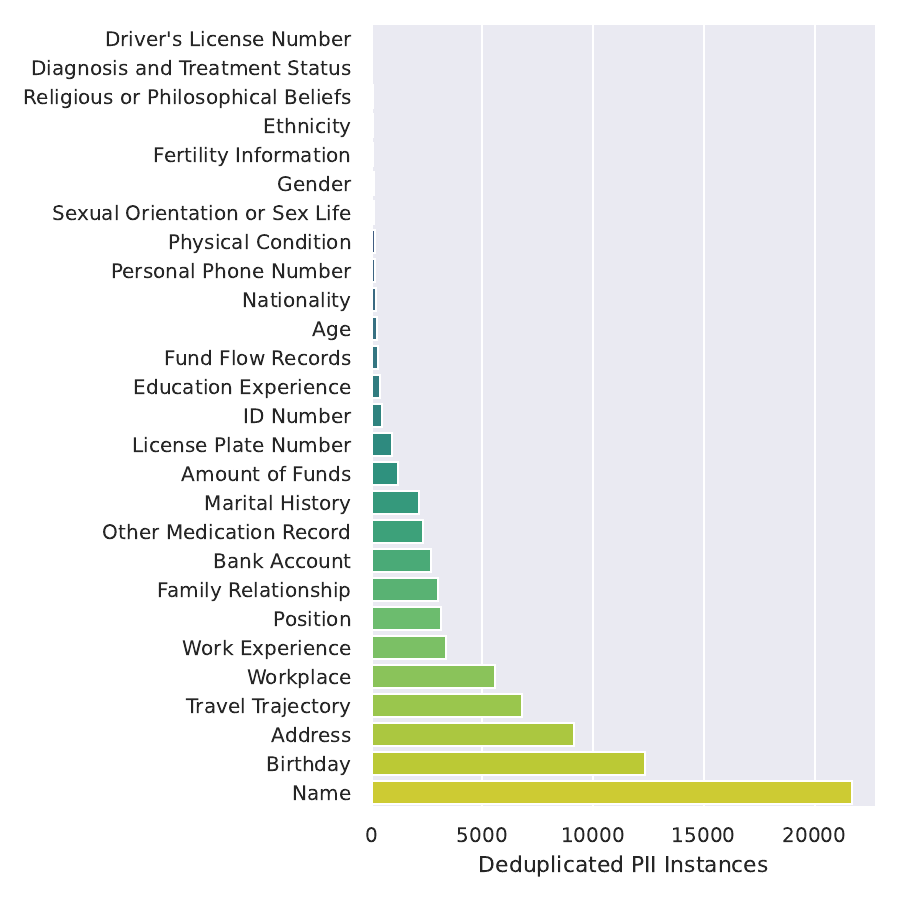}
    \caption{Distribution of de-duplicated PII instances by label category.}
    % \vspace{-1em}
    \label{fig:ds_pii_label_dist}
\end{figure}

We reviewed the definitions and examples of PII in various legal provisions, including CPIS, GDPR, CCPA, and Singapore PDPC (see Acronyms List in Appendix \ref{sec:acronyms}), and used them as references to establish a systematic PII labeling standard. We selected PII types relevant to the text modality and removed types that are unlikely to appear in legal texts (e.g., browser history, SMS content, IP \& MAC addresses), as well as those that are difficult to describe or evaluate (e.g., medical examination reports, psychological trends). Ultimately, we defined labeling guidelines encompassing 7 major categories and 36 subcategories. 
The distribution of labeled PII types is shown in Figure~\ref{fig:ds_pii_label_dist} in the main text, 
whereas a complete summary of these standards is provided in Table~\ref{tab:PII_standards} in Appendix~\ref{app:pii_labeling_standards}.

We employed a combination of machine-assisted annotation and manual verification to label the data. For each major PII category, we designed a dedicated prompt depicted in Figure~\ref{fig:PII_prompt_template} and employed GPT-4o~\citep{openai2024gpt4ocard} to generate annotations. We then recruited students to verify a subset of those annotations with the help of Label Studio~\citep{Label_Studio}. The agreement between human evaluations and GPT-4o annotations achieves an F1 score of 89.9\%. Owing to space limitations, the full evaluation results are shown in Table~\ref{tab:human_evaluation} in Appendix \ref{app:human_annotation_details}, 
together with further details of the annotation process, including annotator backgrounds, annotation instructions, and user interfaces.

\section{Method}

\subsection{Problem Definition}
\label{sec:prob_def}
We study a novel extraction attack tailored to FedLLMs, which differs from traditional verbatim data extraction in three key aspects:

\textbf{Assumptions.} Unlike VDE that assumes the attacker has access to most or all of the training data, our setting limits the attacker to a small, isolated subset of the overall training corpus.

\textbf{Setup.} In our formulation, the prefix and its corresponding target suffix are not drawn from a contiguous span of training data. Instead, extraction prefixes are sampled from the attacker’s local dataset $D_a$, while the target suffixes reside exclusively in other clients’ private data and are absent from $D_a$. Thus, each prefix must generalize beyond local context to trigger the generation of unseen suffixes.

\textbf{Goals.} An attacker does not aim to recover all training completions, but instead focuses on extracting specific, high-value information—most notably, PII—from the global model.

Formally, we consider a \textbf{cross-silo} FL system comprising $c$ clients $\mathcal{C} = \{ C_1, C_2, \dots, C_c \}$, where each client $C_i$ holds a local dataset $D_i$. Among them, one client—denoted as the attacker $C_a \in \mathcal{C}$—is assumed to be \textbf{semi-honest}~\citep{Applebaum2017, source_sia_mia}. That is, $C_a$ faithfully follows the FL protocol (e.g., does not poison data or manipulate model weights), but acts adversarially in a passive manner, attempting to infer PII contained in other clients’ datasets by analyzing the global model $\theta$.

In this setting, the attacker issues queries to the model $\theta$ to extract data without knowing which client any particular output originates from. However, for evaluation purposes, we designate one client as the reference victim to measuring the attack’s effectiveness. Let $S_a$ and $S_v$ denote the sets of PII instances held by the attacker and the victim client, respectively. The attacker constructs a prompt set $P$ and queries the FedLLM $\theta$, obtaining a corresponding output set $Y$. We formalize key definitions and evaluation metrics in the following subsections.

\begin{definition}[Extracted]
A PII instance $s \in S_v$ is considered successfully \emph{extracted} if there exists a prompt $p \in P$ and a corresponding model output $y \in Y$ such that:
\begin{align}
    \exists u \in \Sigma^* \text{ such that } y = s \oplus u,
\label{eq:def_extracted}
\end{align}
where $\Sigma^*$ is the set of all finite-length strings over the vocabulary, and $\oplus$ denotes string concatenation. In other words, the model output $y$ begins with $s$.
\end{definition}

\begin{definition}[Coverage Rate]
The coverage rate measures how thoroughly the attacker recovers the PII unique to the victim client. It is defined as:
\begin{gather}
    S_E = \{ s_i \mid \exists y \in Y \text{ such that } s_i \text{ is extracted by } y \},\nonumber \\
    \text{CR} = \frac{| (S_v \setminus S_a) \cap S_E |}{| S_v \setminus S_a |}. \label{eq:def_CR}
\end{gather}
A higher CR indicates that a larger fraction of the victim’s unique PII has been successfully extracted.
\end{definition}

\begin{definition}[Efficiency]
Efficiency quantifies the precision of extraction with respect to the number of queries. Let $Q$ denote the number of queries, the efficiency is defined as:
\begin{align}
    \text{EF} = \frac{|(S_v \setminus S_a) \cap S_E|}{Q}. \label{eq:def_EF}
\end{align}
A higher EF indicates that more PII is extracted with fewer queries.
\end{definition}

Building upon these definitions, the central challenge is to design algorithms that enable the attacker to extract PII both comprehensively and efficiently—that is, achieving high coverage and high efficiency.

\subsection{Attacking Algorithms}

\subsubsection{PII-contextual Prefix Sampling}
\label{sec:pii_contetual_pref_sampling}
We start with a simple method for constructing query prompts using contextual prefixes of PII, which are word sequences immediately preceding PII instances in the attacker’s dataset $D_i$. This mitigates reliance on manually crafted prompts (e.g., ‘my phone number is’), which often deviate from the model’s training distribution and exhibit limited empirical applicability.

Let the attacker’s training corpus be $U_a = \{t_0, t_1, \dots, t_{|U_a|}\}$, formed by concatenating samples in $D_i$, with $\mathcal{S}$ the multiset of labeled PII. For a PII instance $s \in \mathcal{S}$, let $\text{Loc}(s)$ be the index of its first token in $U_a$. We define a $\lambda$-length contextual prefix function:
$$
\mathcal{T}_\lambda(U, s) = t_{\text{Loc}(s) - \lambda} \cdots t_{\text{Loc}(s) - 1}.
$$
The contextual prefix set of a PII set $\mathcal{S}$ is given by:
\begin{align}
    P_c = \{ \mathcal{T}_\lambda(U_a, s) \mid s \in \mathcal{S} \}.
\end{align}

For each $p \in P_c$, the attacker $\mathcal{C}_a$ queries the global model $\theta$ to generate a suffix $y$ of up to $m$ tokens:
$$
y = \{x_1, \dots, x_m\} \sim \mathbf{P}(y \mid p; \theta).
$$
To enhance diversity, $n$ independent suffixes may be sampled per prefix:
$$
Y_p = \{ y_1, \dots, y_n \}, \quad 
Y = \bigcup_{p \in P_c} Y_p, \quad
Q = n \cdot |P_c|.
$$

A generalized version extends $P_c$ by including all substrings ending before each PII:
$$
\text{SUP}(P_c)= \{ t_i \cdots t_{\text{Loc}(s) - 1} \mid (\text{Loc}(s)-i)\in [1,\lambda] \}.
$$
This yields broader coverage but incurs high query cost due to the large prefix set.

\subsubsection{Frequency-Prioritized Prefix Sampling}
\label{sec:fp_prefix_sampling}

Following prior work~\citep{associative_memorization_shao-etal-2024-quantifying}, which associates extraction effectiveness with co-occurrence frequency, we posit that prefixes frequently occurring before PII entities tend to capture stronger and more diverse associations. Therefore, we emphasize high-frequency prefixes to construct a compact, information-rich set.

Formally, we partition $\text{SUP}(P_c)$ by prefix frequency. For each $\sigma \geq 1$,
\[
P_\sigma = \{ p \in \text{SUP}(P_c) \mid \text{Count}_{\text{SUP}(P_c)}(p) = \sigma \}.
\]
This yields
\[
\text{Set}(\text{SUP}(P_c)) = \bigcup_{\sigma \geq 1} P_\sigma.
\]
Given a threshold $\sigma_a$, the frequent prefix set is
\[
P_{f\geq \sigma_a} = \bigcup_{\sigma \geq \sigma_a} P_\sigma,
\]
sorted by frequency. Setting $\sigma_a=1$ recovers the full contextual prefix set. With a budget $B$, we select the top-$B$ prefixes from $P_{f\geq \sigma_a}$, thereby emphasizing frequent contexts.

\subsubsection{Latent Association Fine-tuning}
We conjecture that a model’s vulnerability to PII extraction stems from its capacity to capture the conditional probability $\text{P}(\mathcal{B} \mid \mathcal{A}; \theta)$ under model parameters $\theta$, where $\mathcal{A}$ denotes prefixes that typically precede PIIs and $\mathcal{B}$ represents the corresponding PII instances.

Since the association between PII and their prefixes is implicitly encoded in the model’s representations, we propose Latent Association Fine-tuning (LAFt), which updates parameters to maximize $\text{P}(\mathcal{B} \mid \mathcal{A}; \theta)$. The goal is to strengthen the mapping between indicative prefixes and PII, thereby improving extraction.

As the first step, we build a fine-tuning dataset $D_{\text{ft}}$ by pairing frequent prefixes with known PII:
\begin{align}
D_{\text{ft}} = \{ (p, s) \mid p \in P_f,\ s \in S_a \},
\end{align}
where $P_f$ is the frequent-prefix set from $D_a$, and $S_a$ the attacker’s PII set. The model is then fine-tuned with the standard causal LM objective:
$$
\theta' = \arg\min_{\theta} \sum_{(p, s) \in D_{\text{ft}}} \sum_{t=1}^{|s|} -\log \mathbf{P}(s_t \mid p, s_{<t};\, \theta).
$$
The updated model $\theta'$ is subsequently used for extraction with prefixes from $P_f$ or $P_c$. 

To note that, LAFt follows the \textbf{semi-honest} FL setting such that the fine-tuned model remains local and is not uploaded to the server.

\section{Experiment}

\subsection{Experimental Setup}
% We detail the experimental setup and implementation to ensure reproducibility.
\paragraph{Federated Setup.}
Our federated setup consists of two main components: the data partitioning across clients and the federated fine-tuning procedure.

For the data partitioning, we simulate a system with 5 clients using a label-skewed non-IID partitioning based on clustering of language embeddings~\citep{ds_clustering}, and ensured that each client receives a comparable number of samples.

For the federated fine-tuning, we perform training on legal tasks using the OpenFedLLM framework~\citep{ye2024openfedllm}, with FedAVG~\citep{fedavg_2017} as the aggregation method over 10 communication rounds. All clients adopt parameter-efficient fine-tuning (LoRA) and a shared prompt template. Hyperparameter settings and implementation details are provided in Appendix~\ref{sec:impl_details}.

After the federated fine-tuning, we evaluate the utility of the final global model on a held-out global test set. Following common practice, we compare it to a centrally trained (non-FL) baseline evaluated on the same test set. The results are reported in Table~\ref{tab:fl_finetuning_performance} in the Appendix.

\paragraph{Models and Metrics.}
Since our data and tasks are derived from Chinese legal documents, we focus primarily on LLMs with proficiency in Chinese. Specifically, we evaluate Qwen1-8B~\citep{bai2023qwen}, Baichuan2-7B~\citep{yang2023baichuan}, Qwen3-8B~\citep{yang2025qwen3technicalreport}, GLM4-8B~\citep{glm2024chatglm}, and Llama3-Chinese~\citep{llama3_chinese_repo}.\footnote{All models are publicly available on HuggingFace: \href{https://huggingface.co/Qwen/Qwen-1_8B}{Qwen1}, \href{https://huggingface.co/baichuan-inc/Baichuan2-7B-Base}{Baichuan2}, \href{https://huggingface.co/Qwen/Qwen3-8B-Base}{Qwen3}, \href{https://huggingface.co/zai-org/GLM-4-9B-0414}{GLM4}, \href{https://huggingface.co/FlagAlpha/Llama3-Chinese-8B-Instruct}{Llama3-Chinese}.}

We evaluate model performance using two primary metrics: Coverage Rate (\textbf{CR}), Efficiency (\textbf{EF}). In addition, we introduce an intermediate metric, Victim-exclusive Extracted PII (\textbf{VxPII}), defined as $|(S_v \setminus S_a) \cap S_E|$, which directly measures the amount of extracted information.

\paragraph{Attack Strategies.}
We designate client 0 as the attacker and client 1 as the victim, and evaluate three strategies:
(1) PII-contextual prefix sampling. The attacker builds a prefix set $P_c$ from its local dataset $D_0$ with prefix length $\lambda=50$. Each prefix queries the global model 15 times, generating up to $m=10$ tokens per query—sufficient to recover most PIIs with manageable cost.
(2) Frequency-prioritized sampling. Prefixes in $\text{Set}(\text{SUP}(P_c))$ are ranked by frequency to form $P_{f\geq 1}$ and used in descending order. Sweeping the prefix budget $B$ varies the frequency threshold $\sigma_a$, enabling analysis of coverage–efficiency trade-offs.
(3) Latent association fine-tuning. The attacker fine-tunes the global model (1 epoch, LR = 5e-5, LoRA: $r=16$, $\alpha=32$) using 10k frequent prefixes and 10k randomly sampled PIIs from its own data to reinforce prefix–PII associations.
Further implementation details are provided in Appendix~\ref{sec:pii_details}.

\paragraph{Evaluation Protocol.}
To ensure a fair evaluation, the set of victim-exclusive PIIs $(S_v \setminus S_a)$ is obtained by applying two filters:
(1) retain only those victim PIIs that do not appear in the attacker’s training corpus (i.e., $s_i \in S_v$ but $s_i \notin U_a$); and
(2) remove PIIs that share a common prefix to avoid ambiguity in identifying which PII was extracted (see Equation~(\ref{eq:def_extracted})).
This is enforced by constraining the length of the longest common prefix (LCP) between any two PIIs:
\[
    \text{LCP}(s_i, s_j) = 0, \quad \forall s_i \neq s_j \in S_v
\]
Metrics defined in Equations~(\ref{eq:def_CR}) and (\ref{eq:def_EF}) are then computed on this filtered, prefix-disjoint set.

\subsection{Results and Discussions}

\paragraph{RQ1: How effective is the PII extraction attack using contextual prefixes?}

We first evaluate the coverage rate (CR) and efficiency (EF) of our extraction attacks by querying federated fine-tuned LLMs using the PII-contextual prefix set $P_c$. Table~\ref{tab:max_cr_result} presents the results. With $P_c$, our attack achieves a considerable CR of 22.93\% on Qwen1-8B and 28.95\% on Baichuan2-7B.

\begin{table*}[t]
\centering
\footnotesize
\caption{
Summary of attack results using the PII-contextual prefix sampling method (with and without LAFt), where client 0 (attacker) targets client 1 (victim). The victim-exclusive set $(S_v \setminus S_a)$ includes 8{,}870 unique PII items.
%Additional statistics used to compute CR and EF are available in Appendix Table~\ref{tab:detailed_max_cr_result}.
}
\label{tab:max_cr_result}

\begin{tabular}{lccccc}
\toprule
\multicolumn{1}{c}{\textbf{Model}} & \textbf{Prefix Set}               & \textbf{CR} & \textbf{EF} & \multicolumn{1}{l}{\textbf{VxPII Count}} & \multicolumn{1}{l}{\textbf{Prefix Set Size}} \\ \midrule
\multicolumn{6}{c}{\textit{\textbf{wo LAFt}}}                                                                                                                                                \\ [0.4em]
Qwen1-8B                           & $P_c$                             & 22.93\%     & 0.1910\%    & 2034                                     & 71006                                        \\
Baichuan2-7B                       & $P_c$                             & 28.95\%     & 0.2411\%    & 2568                                     & 71006                                        \\
Qwen3-8B                           & $P_c$                             & 30.69\%     & 0.2556\%    & 2722                                     & 71006                                        \\
GLM4-9B                            & $P_c$                             & 28.20\%     & 0.2348\%    & 2501                                     & 71006                                        \\
Llama3-Chinese-8B                  & $P_c$                             & 19.73\%     & 0.1643\%    & 1750                                     & 71006                                        \\
Qwen1-8B                           & $\mathrm{Set}(\mathrm{SUP}(P_c))$ & 56.20\%     & 0.0110\%    & 4985                                     & 3013161                                      \\
Baichuan2-7B                       & $\mathrm{Set}(\mathrm{SUP}(P_c))$ & 53.56\%     & 0.0105\%    & 4751                                     & 3013161                                      \\ \midrule
\multicolumn{6}{c}{\textit{\textbf{w LAFt}}}                                                                                                                                                 \\ [0.4em]
Qwen1-8B                           & $P_c$                             & 28.30\%     & 0.2357\%    & 2510                                     & 71006                                        \\
Baichuan2-7B                       & $P_c$                             & 28.46\%     & 0.2370\%    & 2524                                     & 71006                                        \\
Qwen1-8B                           & $\mathrm{Set}(\mathrm{SUP}(P_c))$ & 56.57\%     & 0.0111\%    & 5018                                     & 3013161                                      \\
Baichuan2-7B                       & $\mathrm{Set}(\mathrm{SUP}(P_c))$ & 52.16\%     & 0.0102\%    & 4627                                     & 3013161                                      \\ \bottomrule
\end{tabular}%

% \vspace{-1em}
\end{table*}

\begin{figure}[t]
    \centering
    \hspace{3em}
    \includegraphics[width=1\linewidth]{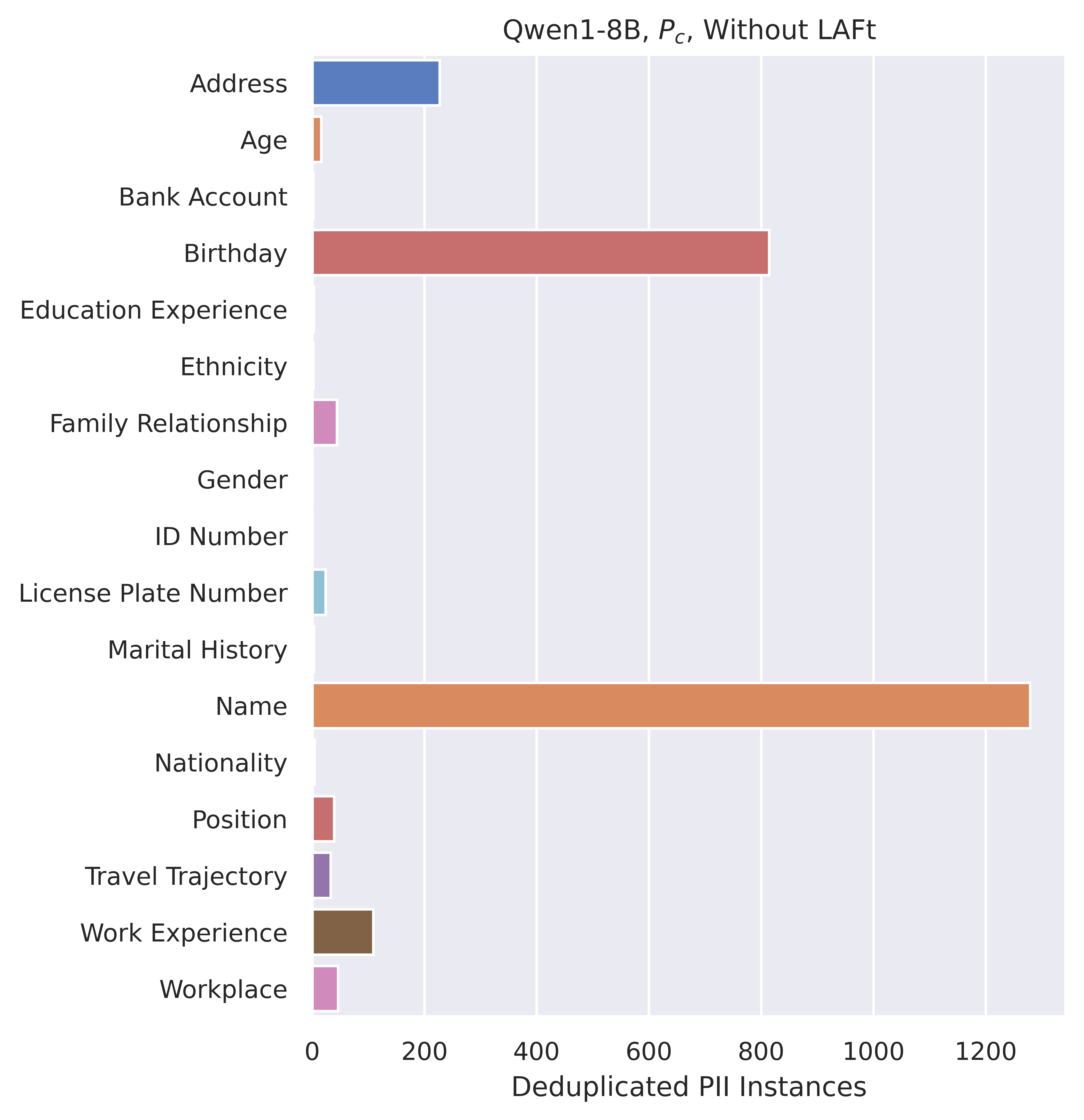}
    \caption{Label distribution of deduplicated victim-exclusive PII extracted by Qwen1-8B (without LAFt, using prefix set $P_c$). Results for Baichuan2-7B are shown in Appendix Figure~\ref{fig:VxPII_distribution_Baichuan2}.}
    \label{fig:VxPII_distribution}
    % \vspace{-1em}
\end{figure}

To understand what types of PII are most vulnerable, we analyze the extracted instances. Figure~\ref{fig:VxPII_distribution} shows the label distribution of deduplicated victim-exclusive PII extracted by Qwen1-8B (without LAFt). The results for Baichuan2-7B are provided in Appendix Figure~\ref{fig:VxPII_distribution_Baichuan2}.

The most frequently extracted PII categories include "Address", "Birthday", and "Name", while others such as "Work Experience" and "Work Place" occur less often but remain notable. More complex types like "Medication Record" are not extracted at all. This is primarily due to the evaluation protocol, which only credits model outputs that match ground truth exactly. Complex PII often appears as long free-text spans, making verbatim reproduction difficult.

\begin{figure}[t]
    \centering
    \includegraphics[width=1\linewidth]{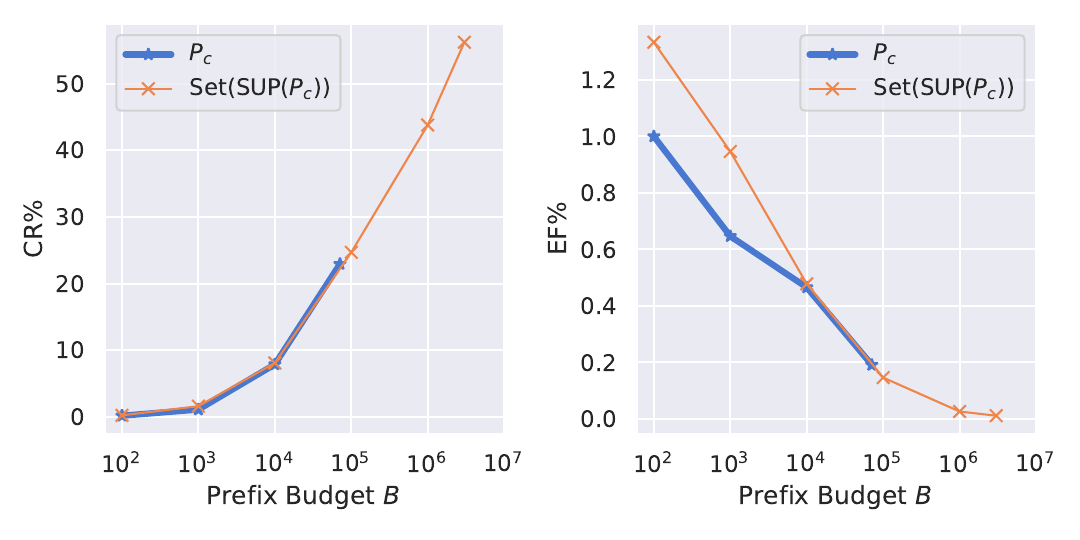}
    \caption{Coverage rate (CR) and efficiency (EF) under varying prefix budgets $B$ for prefix sets $P_c$ and $P_{f \geq 1}$. Prefix set $P_{f \geq 1}$ is frequency-sorted in descending order (see Section~\ref{sec:fp_prefix_sampling}). Budget values are scaled exponentially (base 10); model used is Qwen1-8B.}
    % \vspace{-1em}
    \label{fig:cr_ef_tradeoff_by_prefix_budget}
\end{figure}

To estimate an upper bound of extraction capability, we evaluate with the generalized prefix set $\mathrm{Set}(\mathrm{SUP}(P_c))$, which includes all potential contextual prefixes. As shown in Table~\ref{tab:max_cr_result}, expanding $P_c$ to $\mathrm{Set}(\mathrm{SUP}(P_c))$ increases CR to 56.57\% (Qwen1-8B) and 53.56\% (Baichuan2-7B). However, this gain comes at a steep cost in efficiency—dropping EF to only 0.01\%—indicating most queries yield redundant or irrelevant content.

We further investigate this CR–EF tradeoff in Figure~\ref{fig:cr_ef_tradeoff_by_prefix_budget}, which illustrates how CR and EF vary with prefix budget $B$ for prefix sets $P_c$ and $P_{f \geq 1}$. As $B$ increases, CR improves, but EF declines sharply. This suggests diminishing returns in efficiency when scaling up the number of queries to discover new PII instances.

\paragraph{RQ2: How effective is frequency-prioritized prefix sampling?}

\begin{figure}[t]
    \centering
    \includegraphics[width=1\linewidth]{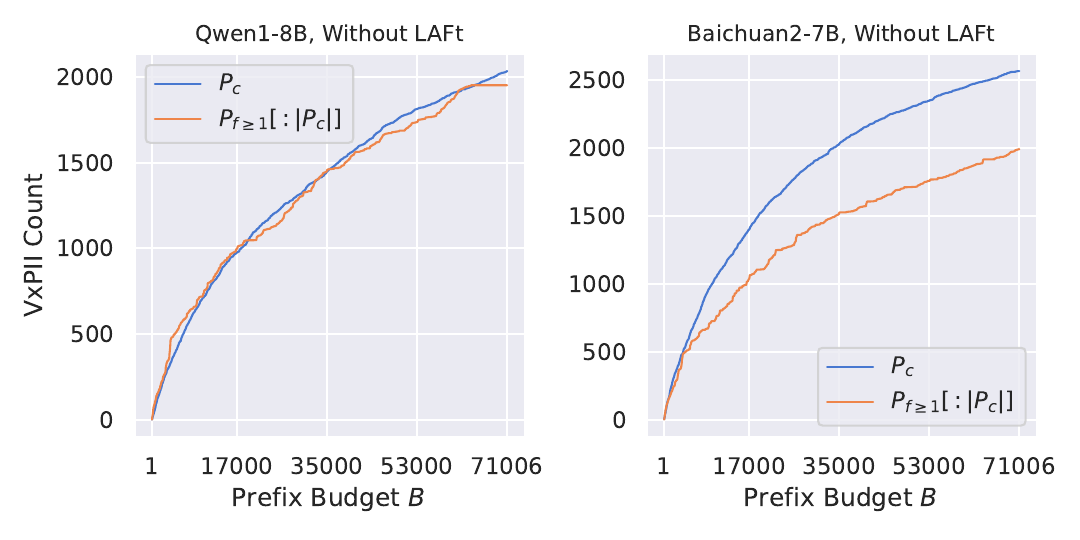}
    \caption{VxPII counts under varying prefix budgets ($B$) for prefix sets $P_c$ and $P_{f \geq 1}$. Prefix set $P_{f \geq 1}$ is frequency-sorted in descending order (see Section~\ref{sec:fp_prefix_sampling}) and truncated to match the size of $P_c$ here.}
    % \vspace{-1em}
    \label{fig:fp_vxpii_cnt_compare}
\end{figure}

As shown in Figure~\ref{fig:fp_vxpii_cnt_compare}, frequency-prioritized (FP) sampling does not extract more VxPII instances than the contextual prefix set $P_c$, contrary to our hypothesis in Section~\ref{sec:fp_prefix_sampling}. This result suggests that the contextual cues embedded in $P_c$ are already strong indicators of LLM memorization, and that memorization cannot be inferred solely from co-occurrence frequency. Instead, it likely arises from more complex interactions between corpus semantics, model architecture, and pre-training dynamics.

\begin{figure}[t]
    \centering
    \footnotesize
    \includegraphics[width=1\linewidth]{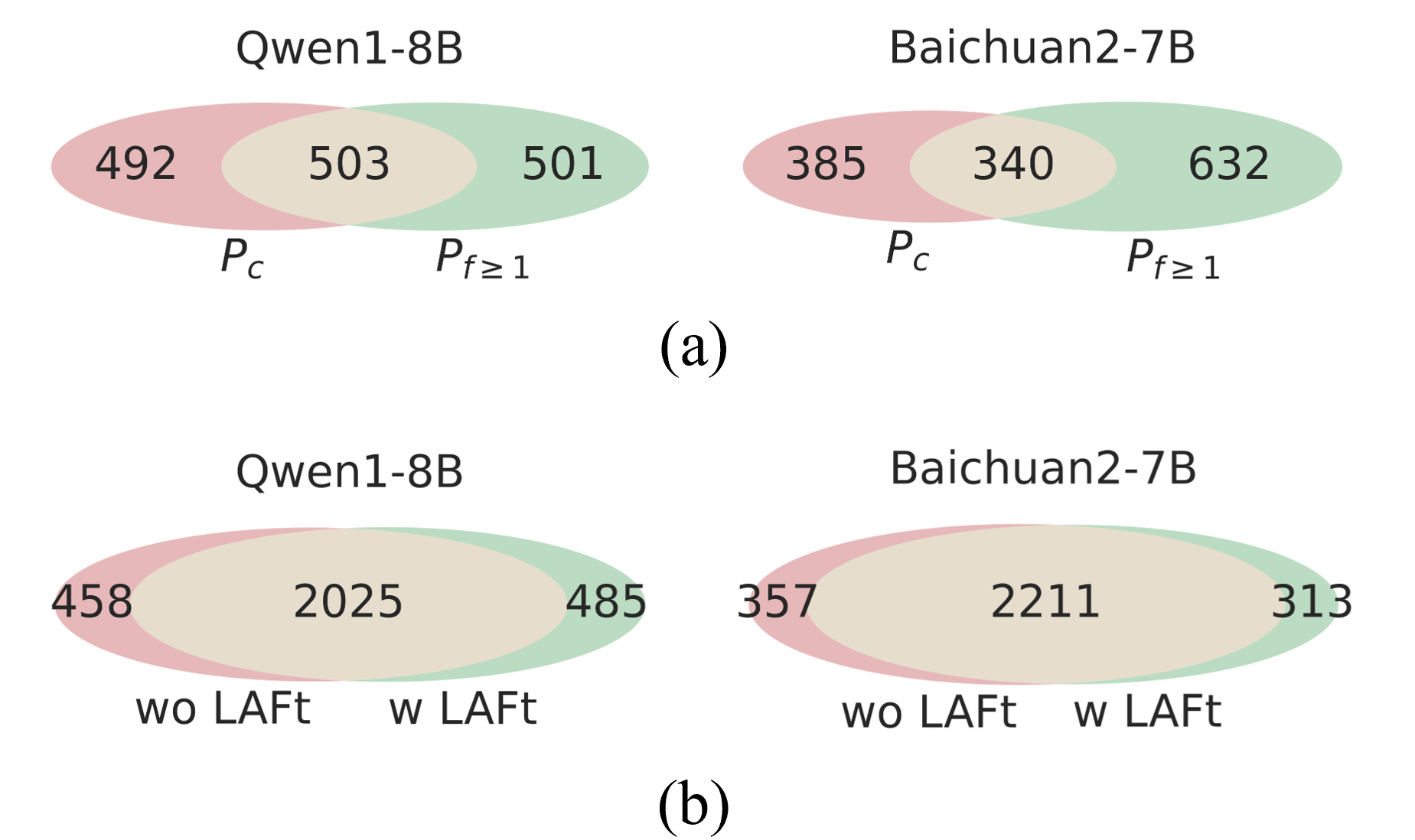}
    \caption{
Venn diagrams showing overlap between VxPII sets extracted by different methods.  
(a) Comparison of VxPII sets using PII-contextual prefixes $P_c$ vs. frequency-prioritized prefixes $P_{f \geq 1}$ at prefix budget $B=10{,}000$ (without LAFt).  
(b) Comparison of VxPII sets extracted with and without LAFt on Qwen1-8B and Baichuan2-7B using the full $P_c$ prefix set.
    }
    % \vspace{-1em}
    \label{fig:venn_method_comparison}
\end{figure}

Despite this, FP sampling captures highly distinct subsets of memorized PII. As shown in Figure~\ref{fig:venn_method_comparison}(a), the Venn diagram comparison reveals that 49.9\% of the VxPII extracted by FP sampling on Qwen1-8B and 65.02\% on Baichuan2-7B are not discovered by the $P_c$ method. This highlights FP sampling’s complementary strength in uncovering diverse memorized content.

\paragraph{RQ3: How effective is Latent Association Fine-tuning?}

As shown in Table~\ref{tab:max_cr_result}, applying Latent Association Fine-tuning (LAFt) significantly improves the CR of Qwen1-8B by 5.37\%, raising it to 28.30\%, and increases EF to 0.24\%, indicating enhanced extraction performance. For Baichuan2-7B, LAFt does not yield a direct improvement in CR, but, as depicted in Figure~\ref{fig:venn_method_comparison}(b), it facilitates the identification of additional distinct PII instances.

These results demonstrate that LAFt is an effective method for increasing the diversity of extracted PII, complementing the FP sampling approach. The extent of the improvement achieved by LAFt is influenced by the construction of the fine-tuning dataset $D_{\text{ft}}$ and the choice of hyperparameters. In this study, we adopt a consistent setting by constructing $D_{\text{ft}}$ through pairing frequent prefixes with randomly sampled PII and fine-tuning the model for one epoch to ensure a fair comparison. However, further exploration of personalized strategies—tailored to models with different architectures and pre-training conditions—could potentially yield better performance.

\subsection{Cross-Client Evaluation of Extraction Robustness}
To assess the robustness of our PII extraction method across different clients, we perform a cross-client evaluation where each client is iteratively designated as the attacker, while the remaining clients act as victims. This setup ensures that the extraction performance is not biased toward any particular client. 

As shown in Table~\ref{tab:res_ablation_cross_client}, our method achieves consistently high coverage rates across all attacker–victim pairings, demonstrating its generalizability and effectiveness in diverse settings.

\begin{table}[t]
\centering
\footnotesize
\setlength{\tabcolsep}{3pt}  % 缩小单元格间距
\caption{Coverage rates (CR) of extraction attacks across different attacker–victim client pairs with a prefix budget $B=10000$. Prefixes are randomly sampled from each attacker's corresponding set $P_c$. “–” indicates self-attack scenarios, which are not applicable.}
\label{tab:res_ablation_cross_client}
\begin{tabular}{@{}cccccc@{}}
\toprule
\multirow{2}{*}{\textbf{\begin{tabular}[c]{@{}c@{}}Attacker\\ ID\end{tabular}}} & \multicolumn{5}{c}{\textbf{Victim ID}}          \\ \cmidrule(l){2-6} 
                                                                                & 0       & 1       & 2       & 3       & 4       \\ \midrule
0                                                                               & -       & 10.91\% & 12.89\% & 10.93\% & 11.88\% \\
1                                                                               & 11.97\% & -       & 12.41\% & 11.46\% & 12.35\% \\
2                                                                               & 12.56\% & 11.39\% & -       & 11.65\% & 12.74\% \\
3                                                                               & 12.07\% & 10.82\% & 12.04\% & -       & 11.99\% \\
4                                                                               & 12.26\% & 11.36\% & 13.25\% & 11.21\% & -       \\ \bottomrule
\end{tabular}
% \vspace{-1em}
\end{table}

% \vspace{-0.5em}

\subsection{PII Sanitization Defense}
\label{sec:masking_defense_experiment}
We evaluate the effectiveness of a simple data sanitization strategy that masks PII using existing annotations. Each PII instance in the training data is replaced with an equal-length sequence of asterisks (*). We then re-fine-tune FedLLM on this sanitized dataset and reapply the PII-contextual Prefix Sampling attack. Table~\ref{tab:res_pii_masking} compares the attack performance with and without the PII masking defense.

\begin{table}[t]
\centering
\footnotesize
\caption{Attack performance with and without PII masking using the contextual prefix set $P_c$. The model is Qwen1-8B.}
\label{tab:res_pii_masking}
\begin{tabular}{@{}lcccc@{}}
\toprule
\multicolumn{1}{c}{}  & \textbf{VxPII} & \textbf{CR} & \textbf{EF} \\ \midrule
With PII Masking                    & 2017           & 22.74\%     & 0.1894\%    \\
Without Defense                     & 2034           & 22.93\%     & 0.1910\%   \\ \bottomrule
\end{tabular}
% \vspace{-1em}
\end{table}

The results show only a slight reduction in the number of extracted VxPII. To investigate this, we analyze the document frequency of extracted VxPII—that is, how often each appears in the training corpus. Figure~\ref{fig:VxPII_doc_freq_dist} demonstrates that masking substantially lowers the frequency of most VxPII, suggesting that our annotations successfully cover the majority of PII. Interestingly, some VxPII with zero document frequency—absent from the sanitized dataset—were still extracted.

Based on these observations, we attribute the limited effectiveness of masking to two factors. First, pretraining data contamination: our training data, drawn from publicly available legal documents, likely overlaps with the pretraining corpora of models such as Qwen1-8B and Baichuan2-7B. Second, incomplete PII labeling: some PII instances may be missing from annotations, and in practice attackers can redefine PII categories, making exhaustive coverage fundamentally difficult.

% \vspace{-0.5em}

\subsection{Disentangling the Effect of Data Contamination}

% \vspace{-0.5em}

Pretraining data contamination is difficult to avoid, as LLM providers rarely disclose their pretraining corpora. To mitigate this influence, we adopt two strategies:

\paragraph{Using open-source LLMs.} To eliminate interference from pretraining memorization, we use open-source or semi-open-source LLMs that are not pretrained on Chinese legal documents: OLMo2-7B~\citep{olmo20242olmo2furious} and Llama2-7B~\citep{touvron2023llama2openfoundation}. As shown in Table~\ref{tab:non_Chinese_LLM_extraction}, after just one round of federated fine-tuning, the CR exceeded 23\%, achieving performance comparable to or better than Qwen1-8B and Baichuan2-7B. This confirms that our findings are not simply an artifact of pretraining contamination.

\begin{table}[t]
\footnotesize
\centering
\caption{Extraction results on non-Chinese LLMs after a single round of federated fine-tuning.}
\label{tab:non_Chinese_LLM_extraction}
\begin{tabular}{@{}lccc@{}}
\toprule
\multicolumn{1}{c}{\textbf{Model}} & \textbf{Prefix Set} & \textbf{CR} & \textbf{EF} \\ \midrule
OLMo2-1124-7B                      & $P_c$               & 23.04\%     & 0.1919\%    \\
Llama-2-7B                         & $P_c$               & 26.41\%     & 0.2200\%    \\ \bottomrule
\end{tabular}
% \vspace{-1em}
\end{table}

\paragraph{Subtracting contaminated memorization.} For Chinese-proficient LLMs that may contain contamination, we adopt a subtraction-based approach. Specifically, we compare the VxPII extracted from the fine-tuned FedLLM ($F$) with those from its base model ($B$), and compute $F \setminus B$ to isolate PII memorized during federated fine-tuning. Table~\ref{tab:pretrain_interference_elimination} shows that even after subtracting $B$, a substantial number of VxPII remain in $F \setminus B$, confirming memorization during fine-tuning. Furthermore, Figure~\ref{fig:fedllm_exclusive_VxPII_dist} demonstrates that $F \setminus B$ exhibits a distribution of VxPII labels similar to Figure~\ref{fig:VxPII_distribution}, supporting the validity of this analysis.

\begin{table}[t]
\footnotesize
\centering
\caption{Comparison of VxPII sets between attacks on FedLLM and its un-fine-tuned base model.}
\label{tab:pretrain_interference_elimination}
\begin{tabular}{@{}ccccc@{}}
\toprule
\multicolumn{1}{l}{Prefix Set} & \multicolumn{1}{l}{Model} & \multicolumn{1}{l}{$|F \setminus B|$} & \multicolumn{1}{l}{$|B \setminus F|$} & \multicolumn{1}{l}{$|F \cap B|$} \\ \midrule
$P_c$                          & Qwen1                  & 682                                   & 518                                   & 1801                             \\
$P_{f\geq 1}$                  & Qwen1                  & 554                                   & 308                                   & 4611                             \\
$P_c$                          & Baichuan2              & 407                                   & 405                                   & 2161                             \\ \bottomrule
\end{tabular}
% \vspace{-1em}
\end{table}

% \vspace{-0.5em}

\section{Conclusion}
To investigate the privacy risks of data extraction attacks in realistic settings, we introduce a new class of attacks targeting FedLLMs. We extend a legal dataset with systematic PII annotations aligned with major privacy regulations, and evaluate attack performance using two key metrics: coverage rate and efficiency. Extensive experiments demonstrate that certain PII types are highly vulnerable, and our proposed methods can achieve substantial extraction performance. These findings highlight a critical privacy gap in FedLLMs and underscore the urgent need for stronger defense mechanisms in future federated learning systems.

% \newpage
% \clearpage
\section*{Limitations}
This work investigates the privacy risks of FedLLMs using a legal-domain dataset. Future research can extend our proposed methods to other sensitive domains such as healthcare and finance, where privacy concerns are equally critical. Additionally, there is a need for further exploration of defense mechanisms that can preserve the privacy of FedLLMs while maintaining their performance.

\section*{Ethics Statement}
This paper presents PII extraction attacks on federated fine-tuned LLMs to expose potential privacy risks. While designed for research and defense purposes, such methods could be misused to recover sensitive user data in real-world FL systems. We conduct all experiments on legal datasets with anonymized PII, and highlight the need for stronger safeguards in FedLLM deployments.

\section*{Acknowledgements}
We thank Guanzhong Chen and Yukun Zhang for their help with dataset annotation. All annotators were properly compensated for their contributions.

\bibliography{custom}

\appendix

\section{Acronyms List}
\label{sec:acronyms}

\begin{itemize}
  \item \textbf{GDPR} - General Data Protection Regulation~\citep{GDPR}
  \item \textbf{CCPA} - California Consumer Privacy Act~\citep{CCPA}
  \item \textbf{CPIS} - Chinese Information Security Technology: Personal Information Security Specification (GB/T 35273-2020)~\citep{GBT35273-2020}
  \item \textbf{Singapore PDPC} - Personal Data Protection Commission (Singapore)~\citep{SingaporePDPC}
  \item \textbf{Non-IID} - Non-independent and identically distributed
\end{itemize}

\section{Preliminary Knowledge}
\label{sec:preliminary}
\subsection{Data Extraction Attack}
\label{sec:preliminary_DEA}
Early research on training data extraction attacks has broadly categorized them into untargeted and targeted attacks~\citep{google_research_2022, bag_of_tricks_icml_2023}. Untargeted extraction aims to recover any memorized training samples without specifying a target~\citep{PII_leakage_2023_IEEE_SP}, whereas targeted extraction attempts to reconstruct specific training samples, often by providing a known prefix and recovering the remaining content~\citep{carlini2021extractingtrainingdatalarge}. The latter type, often referred to as Verbatim Data Extraction, has become a standard approach for evaluating memorization in LLMs~\citep{quantifying_iclr_2023, CDD_acl_2024} and for detecting potential data contamination~\citep{CDD_acl_2024}. We briefly outline the core methodology of verbatim data extraction below.

Given an LLM $\theta$ and a training dataset $X$, each training sample $x_i \in X$ is partitioned into two segments: a prefix $a_i$ and a suffix $b_i$, such that $x_i = a_i b_i$. The model is then prompted with $a_i$ to generate a completion $g_i$ of the same length as $b_i$. If $g_i$ exactly matches $b_i$, the sample is considered successfully extracted.

In practice, model outputs may not exactly replicate the original suffix but can still be lexically close. To accommodate this, a similarity-based metric such as Edit Distance~\citep{edit_distance_1965} is often employed. A sample is deemed extracted if the similarity score between $g_i$ and $b_i$ exceeds a predefined threshold $t$. By computing this similarity-based extraction score across all samples in a dataset $D$, one can quantify the model's memorization behavior or assess its vulnerability to training data extraction attacks.

\subsection{Federated Learning}
Federated Learning (FL) is a solution to address data isolation issues~\citep{fml_2019}, where a central server and multiple clients collaborate to complete the training process. A key feature of FL is that the training datasets are stored locally on each client and remain invisible to other clients. FL is commonly used in industrial scenarios where each client represents an independent organization, such as hospitals collaborating to train a medical model without combining their datasets due to legal restrictions or business competition. Federated Learning enables the training of stronger models compared to training on data from a single client alone.

Given $c$ clients and their private datasets $D_1, D_2, \dots, D_c$, the federated learning process aims to learn a global model $\theta$ by solving the following optimization problem:
$$
\theta^* = \mathop{\arg\min} \limits_{\theta} \frac{1}{c} \sum_{i=1}^{c} \mathcal{L} (D_i, \theta)
$$
To solve this problem, many federated optimization algorithms have been proposed, such as FedAVG~\citep{fedavg_2017} and FedProx~\citep{fedprox_2020}. Typically, these algorithms consist of two alternating phases: local updating and central aggregation. In the local updating phase, each client independently optimizes the global model using its own dataset. In the central aggregation phase, the server aggregates the models from the clients using an aggregation algorithm, obtaining a global model, which is then sent back to each client for the next round of local updating. A typical procedure of federated learning is illustrated in Algorithm \ref{alg:fl_framework}.

\subsection{Federated Large Language Models (FedLLMs)}
\label{app:background_knowledge_FedLLMs}

In current research at the intersection of federated learning and large language models, the term \emph{FedLLM} predominantly refers to \emph{federated fine-tuning} of pre-trained LLMs~\citep{fedfinetune_NEURIPS2024_1a134b50, FLoRA_NEURIPS2024_28312c94, fedllm_benchmark_NEURIPS2024}, rather than federated pre-training. This focus arises from both practical and technical considerations.

Federated pre-training is rarely necessary, as large-scale pre-training typically relies on publicly available general-purpose corpora (such as web text or encyclopedias) that do not contain sensitive information and therefore do not require federated sharing. Moreover, federated pre-training would entail transmitting extremely large model parameters across institutions, incurring prohibitive communication costs and conflicting with high-efficiency training practices like data parallelism and optimized operators, making deployment infeasible in practice.

By contrast, federated fine-tuning is essential in privacy-sensitive domains. Many applications of LLMs in vertical fields—such as judicial documents, electronic health records, and banking customer data—rely on restricted information that cannot be centralized due to legal or institutional constraints (e.g., the Data Security Law or the Personal Information Protection Law). In these settings, federated fine-tuning allows each institution to adapt a shared pre-trained model locally, achieving a balance between model performance and data privacy. This approach has already demonstrated tangible value across multiple domains.

For example, in the judicial domain~\citep{fedlegal_acl_2023}, courts can fine-tune a common LLM on local case repositories without sharing sensitive records, enabling cross-court tasks such as statute matching and case similarity analysis. In healthcare~\citep{ali2025finetuningfoundationmodelsfederated}, hospitals can locally fine-tune models on specialty data, producing comprehensive medical LLMs capable of supporting structured record processing and disease risk prediction. And in finance~\citep{sha2024researchfinancialfraudalgorithm}, banks can fine-tune models on transaction data to detect fraud and assess credit risk without violating privacy regulations.

\subsection{Preliminary Assessment of Verbatim Data Extraction Risks in FedLLMs}
\label{sec:prelim_verbatim_attack}

To examine the memorization behavior of FedLLMs and evaluate their potential risks of leaking sensitive information, we conduct a preliminary experiment simulating a \textit{verbatim data extraction} attack. The results are referenced in the main paper (Section~\ref{sec:introduction}) to empirically motivate our study.

We adapt the experimental setup from~\citep{CDD_acl_2024} to the federated setting, where an attacker is assumed to possess prefix fragments of the training data from all participating clients and attempts to recover the subsequent suffix tokens. For each training sample, we extract a prefix from the original sequence and query the trained model to generate a continuation. The generated suffix is compared against the ground truth using \textbf{Edit Distance} (ED)~\citep{edit_distance_1965}, where a lower ED indicates stronger memorization. Specifically:
\begin{itemize}
    \item ED = 0 indicates the model has perfectly memorized and reproduced the suffix;
    \item ED values are capped at 50, as we restrict suffixes to a maximum of 50 tokens.
\end{itemize}

We perform the attack on the global models aggregated after 10 rounds of federated training. To ensure a comprehensive assessment, we consider three popular FL algorithms—\textbf{FedAvg}~\citep{fedavg_2017}, \textbf{FedProx}~\citep{fedprox_2020}, and \textbf{Scaffold}~\citep{ICML2020_scaffold}—each under both \textbf{IID} and \textbf{Non-IID} data distributions. Two baseline settings are also included:
\begin{itemize}
    \item \textbf{Centralized}: All client data is pooled and the model is fine-tuned in a conventional non-federated manner;
    \item \textbf{Untrained}: The base model is evaluated without any fine-tuning.
\end{itemize}

Figure~\ref{fig:ed_all_last_round} summarizes the results across five downstream tasks. Our key observations are:

\begin{figure*}[!ht]
    \centering

    \includegraphics[width=1\linewidth]{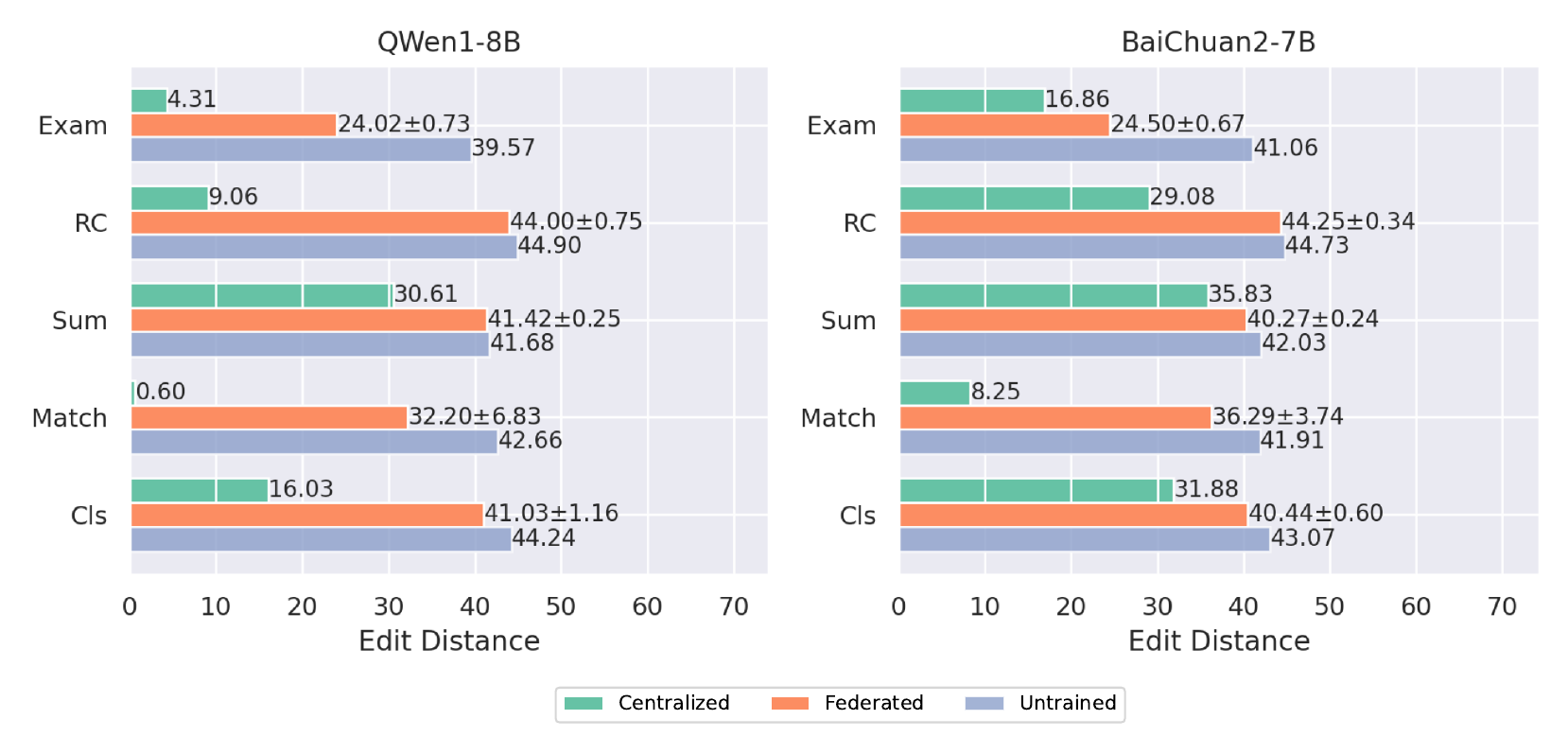}
    
    \caption{Edit Distance results of verbatim data extraction attacks after 10 training rounds. We evaluate six federated configurations (FedAvg/FedProx/Scaffold $\times$ IID/Non-IID) and report the mean and standard deviation. Lower values indicate stronger memorization. Centralized and untrained models serve as baselines.}
    \label{fig:ed_all_last_round}
    % \vspace{-2pt}
\end{figure*}

\begin{itemize}
    \item FedLLMs consistently exhibit higher ED scores (i.e., lower memorization) than centralized models, suggesting that the FL aggregation process reduces susceptibility to verbatim extraction.
    \item However, compared to untrained models, FedLLMs still show non-negligible memorization, with noticeably lower ED scores, indicating partial leakage of training data.
\end{itemize}

These findings highlight a trade-off between collaborative model training and privacy preservation, and they serve as the motivation for our in-depth investigation of privacy risks in FedLLMs.

\section{Federated Learning Framework}

Algorithm~\ref{alg:fl_framework} outlines a general framework for Federated Learning (FL), where a central server coordinates multiple clients to collaboratively train a global model without sharing local data. At each round, the server distributes the current model to all clients, each of which performs local updates based on its private data and sends the updated parameters back. The server then aggregates the received updates to produce a new global model.

\begin{algorithm}[t]
    \footnotesize
    \caption{A Federated Learning Framework}\label{alg:fl_framework}
    \begin{algorithmic}[1]
        \Statex \textbf{Input:} Clients set $\mathcal{C}=\{ c_1,c_2,\dots,c_c \}$ with local datasets $D_1, D_2, \dots, D_c$; total FL rounds $R$; initial global model $\theta_0$; server aggregation function $f_{\text{agg}}$; client loss function $\mathcal{L}$
        \Statex \textbf{Output:} Learned global model $\theta_R$

        \Statex
        \State \textbf{ServerExecute:}
        \For {round $r=1$ to $R$ }
            \For{each client $c_i \in \mathcal{C}$ (in parallel)}
                \State $\theta_r^i \gets$ \Call{ClientUpdate}{$c_i, \theta_{r-1}$}
            \EndFor
            \State $\theta_r \gets f_{\text{agg}}(\{ \theta_r^i \vert c_i \in \mathcal{C} \})$
        \EndFor

        \Statex
        \State \textbf{ClientExecute:}
        \Function{ClientUpdate}{$c_i, \theta_{r-1}$}
            \State $ \theta_r^i \gets \mathop{\arg \min}\limits_{\theta} \mathcal{L}(\theta_{r-1}, D_i)$
            \State \Return $\theta_r^i$
        \EndFunction
    \end{algorithmic}
\end{algorithm}

\section{Task Descriptions and Examples}
\label{sec:tasks_desc}
\begin{enumerate}
\item \textbf{Judicial Summarization (Sum)}:  
The task of judicial summarization aims to extract key information from court judgments and generate concise summaries. The input to this task is a legal document, and the output is a summary of its content. The performance of this task is evaluated using the Rouge-L metric, which effectively measures the similarity between the generated and reference texts based on the longest common subsequence (LCS). Rouge-L is a widely used metric in text generation tasks. In this study, we adopt Rouge-L because it captures both semantic and structural similarities between texts, making it suitable for summarizing judicial documents.

\item \textbf{Judicial Reading Comprehension (RC)}:  
This task focuses on answering legal questions based on court documents to evaluate the model's reading comprehension ability. The input consists of a piece of legal material and a question, and the task requires answering the question based on the content of the material. The performance metric for this task is Rouge-L.

\item \textbf{Similar Case Matching (Match)}:  
In this task, the input includes three case documents, and the model is required to determine which of the latter two documents is more similar to the first one. The model selects the most similar document by computing the similarity between the first case and each of the other two. The evaluation metric for this task is accuracy.

\item \textbf{Judicial Exam (Exam)}:  
This task simulates multiple-choice questions from legal examinations to assess the model’s knowledge of legal concepts. Given a judicial exam question with multiple options, the model is expected to choose the correct answer. The performance is evaluated using accuracy.

\item \textbf{Legal Case Classification (Cls)}:  
This task requires the model to classify the cause of action in a case, assisting legal retrieval systems in automatically categorizing case types. The input is a description of the case facts, and the model is required to output the corresponding case category. The performance metric is accuracy.
\end{enumerate}

\begin{table*}[ht]
\centering
\caption{Input and Output Examples for Each Task}
\label{tab:task_input_output_eg}
\resizebox{\textwidth}{!}{%
\begin{tabular}{@{}lll@{}}
\toprule
\textbf{Task}                                                                 & \textbf{Input}                                                                                                                                                                                                                                                                                                                                       & \textbf{Output}                                                                                                                                                   \\ \midrule
\begin{tabular}[c]{@{}l@{}}Judicial Summarization\\ (SUM)\end{tabular}        & \begin{tabular}[c]{@{}l@{}}First-instance civil judgment on inheritance dispute between Han and Su\\ Shenyang Dadong District People’s Court\\ Plaintiff: Han, female, born June 6, 1927, Han ethnicity...\\ …\\ Clerk: Li Dan\end{tabular}                                                                                                          & \begin{tabular}[c]{@{}l@{}}Summary: This case involves an inheritance\\ dispute between the plaintiff and the defendant.\\ The plaintiff requests...\end{tabular} \\ \midrule
\begin{tabular}[c]{@{}l@{}}Judicial Reading Comprehension\\ (RC)\end{tabular} & \begin{tabular}[c]{@{}l@{}}Case: Upon trial, it was found that on February 11, 2014, the plaintiff...\\ Question: When did the plaintiff and defendant agree on the travel plan?\end{tabular}                                                                                                                                                        & \begin{tabular}[c]{@{}l@{}}The plaintiff and defendant agreed on the \\ travel plan on February 11, 2014.\end{tabular}                                            \\ \midrule
\begin{tabular}[c]{@{}l@{}}Similar Case Matching\\ (Match)\end{tabular}       & \begin{tabular}[c]{@{}l@{}}Determine whether Case A is more similar to Case B or Case C.\\ A: Plaintiff: Zhou Henghai, male, born October 17, 1951...\\ B: Plaintiff: Huang Weiguo, male, Han ethnicity, resident of Zhoushan City...\\ C: Plaintiff: Zhang Huaibin, male, resident of Suzhou City, Anhui Province, Han ethnicity...\end{tabular}    & B                                                                                                                                                                 \\ \midrule
\begin{tabular}[c]{@{}l@{}}Judicial Exam\\ (Exam)\end{tabular}                & \begin{tabular}[c]{@{}l@{}}Wu was lawfully pursued by A and B... Which of the following analyses is correct?\\ A. If Wu missed both A and B, and the bullet...\\ B. If Wu hit A, resulting in A’s death...\\ C. If Wu hit both A and B, causing A's death and B’s serious injury...\\ D. If Wu hit both A and B, causing both to die...\end{tabular} & A                                                                                                                                                                 \\ \midrule
\begin{tabular}[c]{@{}l@{}}Legal Case Classification\\ (Cls)\end{tabular}     & Legal document: Plaintiff Yan Qiang submitted the following claims to this court:...                                                                                                                                                                                                                                                                 & Private Loan Dispute                                                                                                                                              \\ \bottomrule
\end{tabular}%
}
\end{table*}

\section{Data Preprocessing}
\label{sec:ds_preprocessing}
Previous works~\citep{fedlegal_acl_2023, disclawllm_finetuninglargelanguage_yue2023} have used these datasets for LLM and FedLLM research. In this work, we use the processed datasets from these prior studies and further curate the data for our experiments. We applied the following preprocessing steps to prepare the datasets:

\paragraph{Deduplication and Cleansing.} To ensure the quality of our data, we remove duplicate samples with logically equivalent meanings. For example, in the RC tasks, some samples only differ in the order of two legal cases. We also clean out samples containing garbled characters or large segments with a mixture of multiple languages.

\paragraph{Unifying Prompt Template and Instruction Reshaping.} Some tasks, such as Exam, contain instructions that appear in different parts of the sample (either at the beginning or the end). To standardize the format, we reshape the data so that the instruction always appears at the beginning, followed by the legal document. Additionally, we employ hierarchical hyper markers such as "<Case A>", "<Case B>", and "<Answer>" to clearly segment the prompt, making the structure more transparent for the LLM.

\section{Supplementary Dataset Statistics and Analysis}

Table~\ref{tab:dataset_stats} summarizes the basic statistics of the five datasets used in our experiments. Each dataset corresponds to a different downstream task for fine-tuning the model.

\begin{figure}[ht]
    \centering
    \hspace*{-20pt}
    \includegraphics[width=0.9\linewidth]{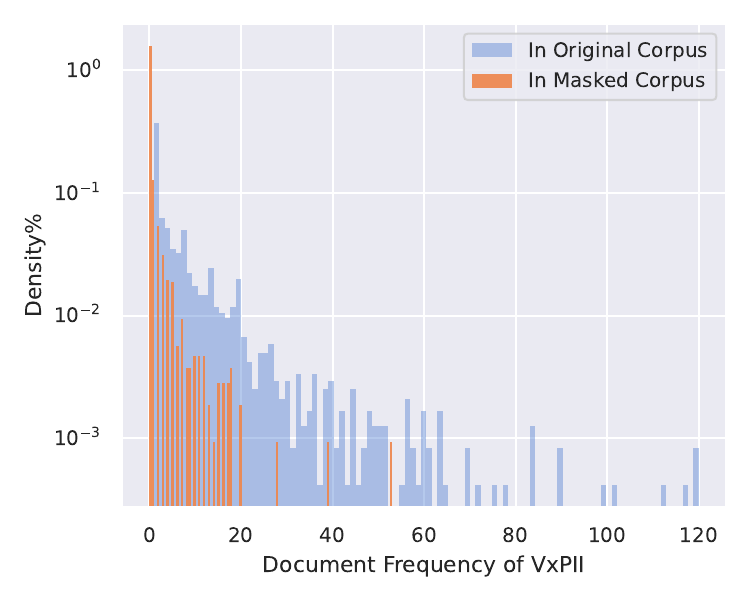}
    \caption{Document frequency distribution of the 2017 VxPII instances extracted from the model trained on the masked dataset.}
    \label{fig:VxPII_doc_freq_dist}
    % \vspace{-1em}
\end{figure}

Figure~\ref{fig:VxPII_doc_freq_dist} presents the document frequency distribution of the 2017 VxPII instances extracted from the model trained on the masked dataset (see Section~\ref{sec:masking_defense_experiment}). Most VxPII exhibit low frequency, indicating that PII masking significantly reduces memorization.

\begin{table}[ht]
    \centering
    \footnotesize
    \caption{Dataset Statistics}
    \label{tab:dataset_stats}

    \begin{tabular}{lccccc}
    \toprule
      & \textbf{Exam} & \textbf{RC} & \textbf{SUM} & \textbf{Match} & \textbf{Cls} \\ \midrule
        \#Samples & 2399          & 3500        & 2651         & 3848           & 4196         \\ \bottomrule
        \end{tabular}

\end{table}

\section{Prompt Template and Utility Fine-tuning Results for FedLLMs}
Figure~\ref{fig:utility_ft_template} shows the unified prompt template used for all federated utility fine-tuning tasks. Table~\ref{tab:fl_finetuning_performance} reports the evaluation results across multiple tasks, comparing different federated learning algorithms and base models.

\begin{figure}[ht]
\centering
\begin{lstlisting}[breaklines=true, basicstyle=\ttfamily\scriptsize,frame=single]
Below is a task related to judicial and legal matters. Output an appropriately completed response to the request.

<### Input>
{{Task Input}}

<### Output>
{{Task Output}}
\end{lstlisting}
\caption{Unified Utility Fine-tuning Template for All Tasks.}
\label{fig:utility_ft_template}
% \vspace{-2em}
\end{figure}

\begin{table*}[htbp]
\centering
\footnotesize
\caption{Utility Performance over Different Tasks.}
\label{tab:fl_finetuning_performance}

\begin{tabular}{@{}ccccccc@{}}
\toprule
\textbf{FL Algorithms} & \textbf{Models} & \textbf{SUM(rouge-l)} & \textbf{RC(rouge-l)} & \textbf{Match(Acc)} & \textbf{Exame(Acc)} & \textbf{Cls(Acc)} \\ \midrule
FedAvg                 & Qwen1-8B        & 50.0                  & 14.2                 & 50.0                & 37.5                & 90.0              \\
FedAvg                 & Baichuan2-7B    & 57.6                  & 42.4                 & 50.0                & 33.3                & 89.5              \\
Non-FL                 & Qwen1-8B        & 50.0                  & 18.9                 & 50.0                & 40.8                & 87.0              \\ \bottomrule
\end{tabular}%

\end{table*}

\section{Additional PII Label Distribution Results}

\begin{figure}[t]
    \centering
    \includegraphics[width=0.8\linewidth]{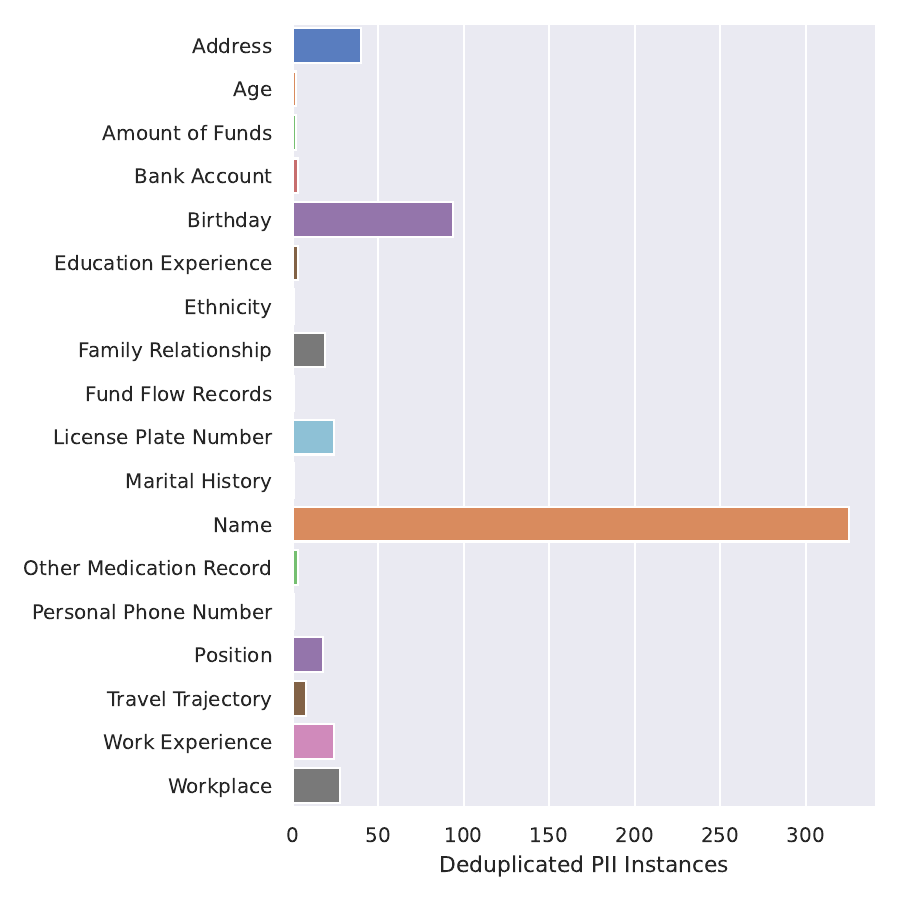}
    \caption{Label distribution of FedLLM-exclusive VxPII extracted using prefix set $P_c$ and the Qwen1-8B model.}
    \label{fig:fedllm_exclusive_VxPII_dist}
\end{figure}

Figure~\ref{fig:fedllm_exclusive_VxPII_dist} illustrates the label distribution of FedLLM-exclusive victim PII extracted by Qwen1-8B. This result corresponds to the experiment described in Section~\ref{sec:masking_defense_experiment}.

\begin{figure}[t]
    \centering
    \includegraphics[width=0.8\linewidth]{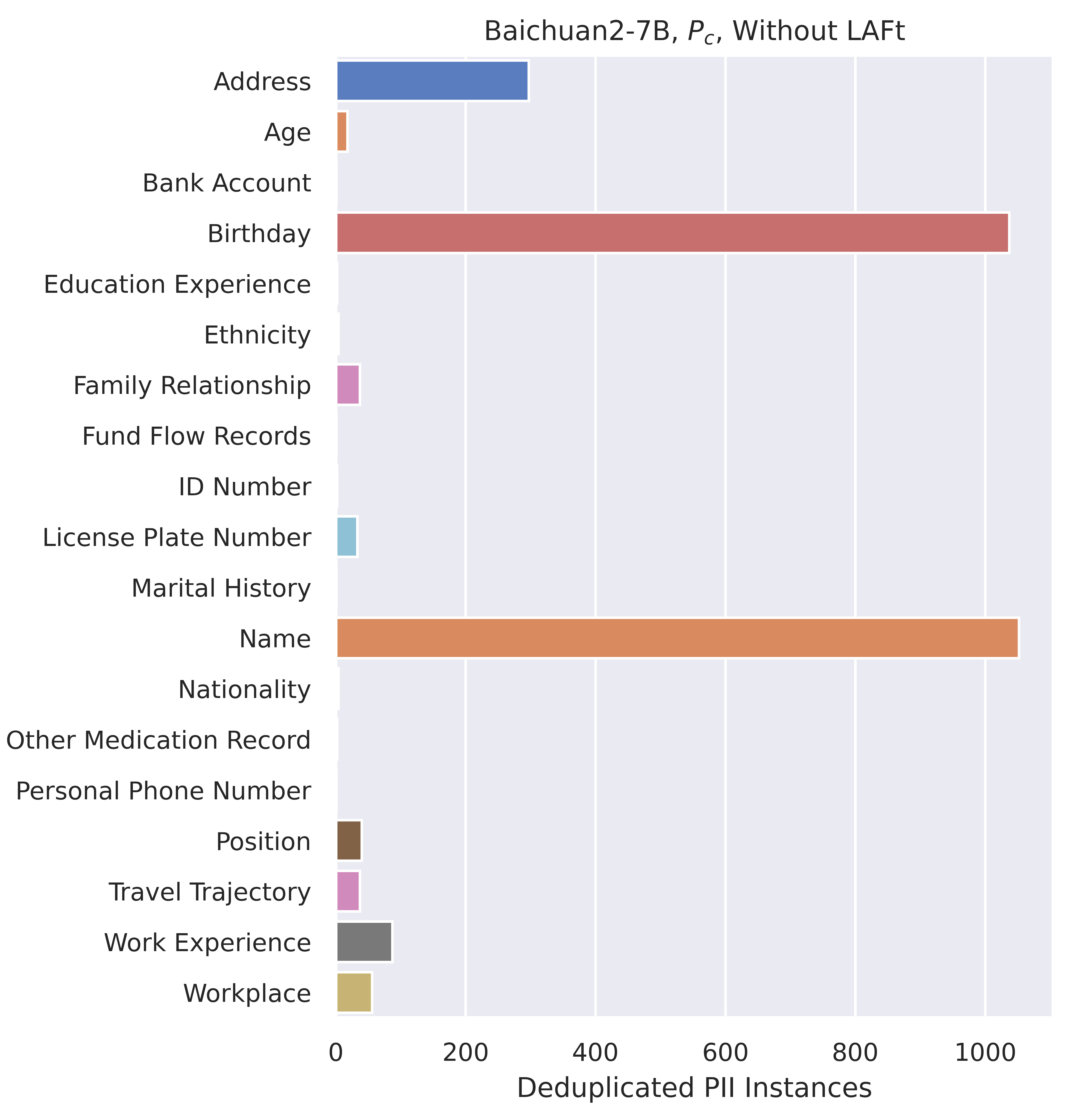}
    \caption{
    Label distribution of deduplicated victim-exclusive PII instances extracted by the Baichuan2-7B model (without LAFt, using prefix set $P_c$). This figure complements Figure~\ref{fig:VxPII_distribution} in the main text, which presents the corresponding results for Qwen1-8B.
    }
    \label{fig:VxPII_distribution_Baichuan2}
    % \vspace{-1em}
\end{figure}

Figure~\ref{fig:VxPII_distribution_Baichuan2} presents the label distribution of deduplicated victim-exclusive PII instances extracted by the Baichuan2-7B model.

\section{Machine Annotation Standards for PII Labeling}
\label{app:pii_labeling_standards}

This section provides details on the machine annotation protocol we use to identify PII in our dataset. Table~\ref{tab:PII_standards} defines our categorization schema, which includes seven major categories and their corresponding fine-grained subtypes. To ensure annotation consistency and scalability, we utilize a templated prompting approach for automated PII labeling. Figure~\ref{fig:PII_prompt_template} shows the machine annotation prompt used to instruct the LLM annotator. The prompt dynamically incorporates category definitions and format constraints to standardize the output.

\renewcommand{\arraystretch}{1.5}
\setlength{\arrayrulewidth}{0.3pt}
\begin{table*}[htbp]
\centering
\footnotesize
\caption{Categorization of PII Types in Our Labeling Standards}
\label{tab:PII_standards}
\begin{tabular}{@{}p{0.3\textwidth} p{0.6\textwidth}@{}}

\toprule
\textbf{Major Category}                 & \textbf{Minor Category}                                                                                  \\ \midrule
Personal Basic Information              & Name, Birthday, Address, Gender, Ethnicity, Family Relationship, Age, Nationality, Personal Phone Number \\
\hline
Personal Identity Information           & ID Number, Social Security Number, Driver's License Number, Employee Number, License Plate Number        \\
\hline
Health Related Information &
  Physical Condition, Fertility Information, Current Medical History, Diagnosis and Treatment Status, Other Medication Record \\
\hline
Work and Education Information & Workplace, Position, Work Experience, Education Experience, Grades                                       \\
\hline
Personal Property Information           & Bank Account, Amount of Funds, Fund Flow Records, Virtual Assets, Other Financial Records                \\
\hline
Personal Location Information           & Precise Location, Accommodation Information, Travel Trajectory                                           \\
\hline
Others &
  Marital History, Religious or Philosophical Beliefs, Sexual Orientation or Sex Life, Unpublished Criminal Records \\ \bottomrule
\end{tabular}
\end{table*}
\renewcommand{\arraystretch}{1.0}

\begin{figure}[ht]
\centering
\begin{lstlisting}[breaklines=true, basicstyle=\ttfamily\scriptsize,frame=single]
I would like you to assist in reviewing the provided document and labeling all sections containing {{Major Categories of PII}} according to the following requirements.

1. **Types of personal information to identify include:**
    {{PII Subcategories}}
2. **Output format:**
    {{Output Format Description}}
3. **Input instructions:**
    {{Input Format Description}}

Please provide the output directly in accordance with the format requirements above, without any additional explanation or comments. Thank you for your assistance!
\end{lstlisting}
\caption{PII Machine Annotation Prompt Template}
\label{fig:PII_prompt_template}
% \vspace{-1em}
\end{figure}

\section{Details of Human Evaluation for PII Annotation}\label{app:human_annotation_details}
\begin{figure*}[ht!]
\centering
\includegraphics[width=1\linewidth]{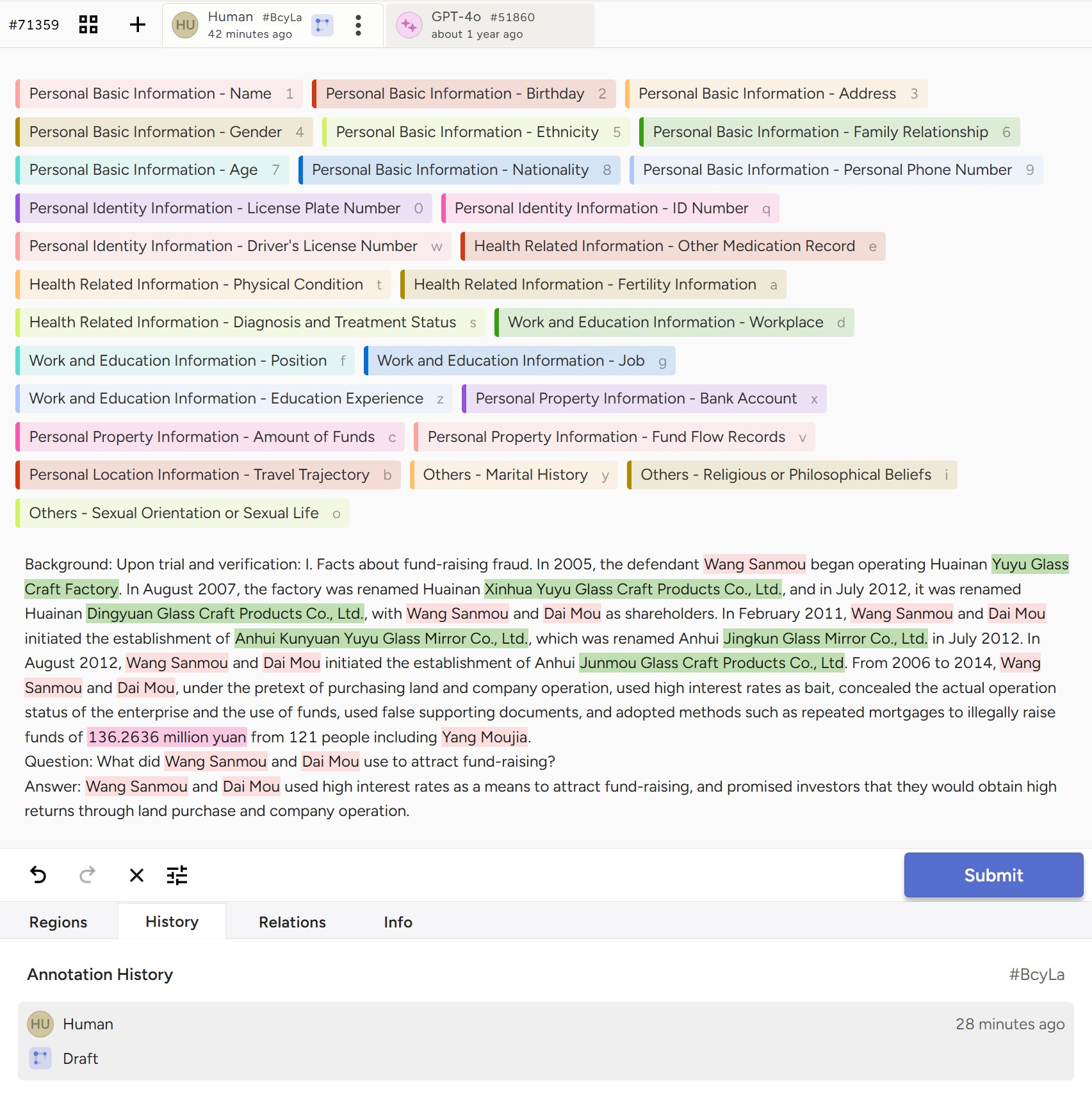}
\caption{Human annotation interface in the Label Studio tool for PII labeling. Annotators are familiar with the Label Studio environment and are instructed to label PII spans based on predefined PII categories. Machine-generated labels are provided as references to assist the human annotators.}
\label{fig:Label-Studio-Human-Interface}
% \vspace{-1em}
\end{figure*}

To validate the quality of the machine-generated PII annotations, we recruited four Chinese-speaking students with foundational knowledge of Chinese law to manually annotate PII on a selected subset of the dataset. Prior to annotation, all annotators underwent thorough training on the annotation guidelines and usage of the Label Studio tool. The instructions provided to annotators are detailed in Figure~\ref{fig:annotation_guidelines}, while Figure~\ref{fig:Label-Studio-Human-Interface} illustrates the annotation interface used. All annotators were fairly compensated upon completion of their tasks.

The human evaluation results, reported in terms of precision, recall, and F1 score, are summarized in Table~\ref{tab:human_evaluation}, indicating high agreement both in exact span matching and in combined span-and-label matching, confirming the reliability of the machine annotations.

\begin{figure*}[htbp]
\centering
\begin{lstlisting}[breaklines=true, basicstyle=\ttfamily\scriptsize,frame=single]
# **PII Annotation Guidelines for Labelers**
## **1. Task Objective**
**Core Task**: Proofread legal texts to accurately identify and annotate **Personally Identifiable Information (PII)**. Each annotation task includes:
1. **Localization**: Mark the exact character offsets of each PII instance in the text;
2. **Categorization**: Assign each PII instance to the appropriate **major category (7 total)** and **minor category (36 total)**, ensuring precise classification.
## **2. PII Category System**
| Major Category | Minor Categories |
| - | - |
| Personal Basic Information | Name, Birthday, Address, Gender, Ethnicity, Family Relationship, Age, Nationality, Personal Phone Number |
...(omitted)...
## **3. Annotation Workflow and Standards**
### **Step-by-Step Process**
1. **Read the Full Text**: Understand the context to detect all potential PII entities;
2. **Sentence-by-Sentence Annotation**: For each PII instance, annotate its **start position**, **text span**, and corresponding **major + minor category**;
3. **Special Cases**: For ambiguous expressions (e.g., "a certain district of a certain city"), determine PII status based on contextual clues.
### **Annotation Guidelines**
* **Accuracy**: Ensure all annotated content is verifiably present in the text. Avoid false positives or over-labeling;
* **Support Channel**: If any uncertain cases arise during annotation, promptly reach out to the *Annotation Support Team* for clarification.

\end{lstlisting}
\caption{Markdown-style guideline for PII annotation, covering task objectives, taxonomy, and labeling procedures.}
\label{fig:annotation_guidelines}
\end{figure*}

\begin{table*}[htbp]
\centering
\footnotesize
\caption{Human Evaluation of PII Labeling Quality}
\label{tab:human_evaluation}
\begin{tabular}{@{}lcccc@{}}
\toprule
\textbf{Evaluation Criteria} & \textbf{Precision (P)} & \textbf{Recall (R)} & \textbf{F1 Score (F1)} \\ \midrule
Identical Span Only           & 0.89                  & 0.93                & 0.91                   \\
Identical Span and Label      & 0.89                  & 0.90                & 0.89                   \\ \bottomrule
\end{tabular}
\end{table*}

\section{Experiment Implementation Details}
\label{sec:impl_details}
\subsection{Federated Dataset Partitioning.} We use the preprocessed and labeled datasets (see Section \ref{sec:dataset}) for our experiments, splitting the data into training and testing sets. In the federated learning setup, we simulate a system with five clients. The testing set remains global, while the training set is heterogeneously partitioned across the clients using a balanced Non-IID distribution (see Acronyms List \ref{sec:acronyms}). To achieve this, We employ a clustering-based method~\citep{ds_clustering} for partitioning, where a language encoder first generates embeddings, which are then clustered using K-means. Finally, a Dirac distribution with $\alpha=0.5$ is applied to create a label-skewed partitioning~\citep{label_skewed_guo2024exploringvacantclasses}, ensuring each client receives a comparable number of samples.

\subsection{Hardware and Computation Budget}
All experiments are conducted on a single NVIDIA A6000 GPU with 48 GB of memory, using bfloat16 precision. Most sampling-based attack experiments are completed within 200 GPU hours.

\subsection{Experiment Procedure}

\subsubsection{Federated Utility Fine-Tuning}
We begin by performing federated fine-tuning of the LLM~\citep{fedlegal_acl_2023, survey_fed_LLM_wu2025surveyfederatedfinetuninglarge} on the partitioned dataset, adapting it to the legal tasks. The fine-tuning is conducted using the OpenFedLLM framework~\citep{ye2024openfedllm}. We set the total number of FL rounds to 10 and use FedAVG~\citep{fedavg_2017} as the aggregation algorithm.

Each client performs multi-task fine-tuning by mixing all local tasks and applying a unified prompt template, as illustrated in Figure~\ref{fig:utility_ft_template}, following the approach of~\citet{t5_raffel2023exploringlimitstransferlearning}. In each round of federated learning, the client fine-tunes the received global model for one epoch using parameter-efficient fine-tuning (PEFT) techniques of LoRA~\citep{lora_hu2021loralowrankadaptationlarge}. The learning rate is set to 3e-4 with a linear decay schedule. The maximum input sequence length is 3072 tokens. We use a batch size of 1 and apply gradient accumulation with a factor of 8. The LoRA configuration is set to $r=16$ and $\alpha=32$.

After federated fine-tuning is complete, we evaluate the utility performance of the final global model on a held-out global test set. In line with standard practices in federated learning research, we also compare this performance with that of a centrally (non-FL) trained model on the same test set. The results are summarized in Table~\ref{tab:fl_finetuning_performance}.

\subsubsection{PII Extraction}
\label{sec:pii_details}
In the main experiments, we designate client 0 as the attacker and client 1 as the victim. We construct the prefix set $P_c$ for PII-contextual prefix sampling from the local dataset $D_0$. During this construction, we set the length parameter $\lambda$ to 50. Each prefix is used to independently query the utility fine-tuned global model $n=15$ times. For each query, the model is allowed to generate up to $m=10$ new tokens. This generation length is sufficient to cover most labeled PII instances while keeping the computational cost acceptable.

For Frequency-Prioritized Prefix Sampling, we construct $\text{Set}(\text{SUP}(P_c))$ from the aforementioned $P_c$, and sort it in descending order of prefix frequency (as described in Section~\ref{sec:pii_contetual_pref_sampling}). The model $\theta$ is then queried using prefixes in this frequency-descending order. Although we do not explicitly define a frequency threshold $\sigma_a$, we sweep the prefix budget $B$ exponentially in base 10. Because $\text{Set}(\text{SUP}(P_c))$ is sorted by decreasing frequency, this sweep over $B$ implicitly corresponds to sweeping $\sigma_a$ from $+\infty$ to 1.

\subsubsection{Latent Association Fine-tuning}
To construct the fine-tuning dataset $D_\text{ft}$, we select the top 10000 most frequent prefixes from $\text{Set}(\text{SUP}(P_c))$ and randomly sample 10000 PII instances from the attacker's (client 0's) PII set $S_a$. Although alternative strategies could be explored for prefix and PII selection, this approach is relatively straightforward and effective. We then fine-tune the model $\theta$ to obtain $\theta'$ using one epoch and a small learning rate of 5e-5. LoRA is applied with $r=16$ and $\alpha=32$, consistent with the initial federated fine-tuning setup.

\end{document}